\documentclass[conference]{IEEEtran}
\IEEEoverridecommandlockouts

\usepackage{times}
\usepackage{latexsym}
\usepackage{graphicx}  
\usepackage{url}
\usepackage{amssymb}
\usepackage{subfigure}
\usepackage{url}
\usepackage{float} 
\usepackage{cite}
\usepackage{amsmath,amssymb,amsfonts}
\usepackage{algorithmic}
\usepackage{textcomp}
\usepackage{xcolor}
\def\BibTeX{{\rm B\kern-.05em{\sc i\kern-.025em b}\kern-.08em
    T\kern-.1667em\lower.7ex\hbox{E}\kern-.125emX}}
\begin{document}

\title{Style-aware Neural Model with Application in Authorship Attribution}

 \author{
\IEEEauthorblockN{Fereshteh Jafariakinabad}
\IEEEauthorblockA{
\textit{Computer Science Department } \\
\textit{ University of Central Florida}\\
fereshteh.jafari@knights.ucf.edu } 
 \and 
 \IEEEauthorblockN{Kien A. Hua}
\IEEEauthorblockA{
\textit{Computer Science Department } \\
\textit{ University of Central Florida}\\
kienhua@cs.ucf.edu  }
}


\maketitle

\begin{abstract}
 Writing style is a combination of consistent decisions associated with a specific author at different levels of language production, including lexical, syntactic, and structural. In this paper we introduce a style-aware neural model to encode document information from three stylistic levels and evaluate it in the domain of authorship attribution. First, we propose a simple way to jointly encode syntactic and lexical representations of sentences. Subsequently, we employ an attention-based hierarchical neural network to encode the syntactic and semantic structure of sentences in documents while rewarding the sentences which contribute more to capturing the writing style. Our experimental results, based on four benchmark datasets, reveal the benefits of encoding document information from all three stylistic levels when compared to the baseline methods in the literature.

\end{abstract}

\begin{IEEEkeywords}
Style-aware Neural Model, Syntax encoding, Authorship Attribution
\end{IEEEkeywords}

\section{Introduction}
Individuals express their thoughts in different ways due to many factors including the conventions of the language, educational background, and the intended audience, etc. In written language, the combination of consistent conscious or unconscious decisions in language production, known as writing style, has been studied widely \cite{koppel2009computational,neal2017surveying}. Stylistic features are generally \textit {content-independent}.  They are consistent across different documents written by a specific author (or author groups). Lexical, syntactic, and structural features are three main families of stylistic features. Lexical features represent author's character and word use preferences, while syntactic features capture the syntactic patterns of the sentences in a document. Structural features reveal information about how an author organizes the sentences in a document.

To date, the existing approaches in the domain of authorship attribution fall into two categories. The first category adopts traditional machine learning techniques to identify the author of a given document. In this approach the stylistic features are engineered and extracted from the documents and are subsequently used as the inputs of different kind of classifiers \cite{segarra2015authorship, castillo2015author, stamatatos2011plagiarism,varela2011selecting, pillay2010authorship, seroussi2011authorship}. These features reveal statistical information of documents in lexical, syntactic, and structural levels. For instance, frequency of certain words, character distribution, function word distribution, frequency of part of speech tags, the number of sentences per paragraph, etc. A limitation of this approach is that the feature extracting process ignores rich sequential information in the sentences and the document.

The second category of authorship attribution approach builds upon neural network models \cite{shrestha2017convolutional,ferracane2017leveraging, hitschler2017authorship}. In this approach, the sequence of words or characters are the input of a neural network which makes the proposed models utilize the sequential information. However, the proposed models in the literature mainly focus on lexical features despite the fact that lexical-based language models have very limited scalability when dealing with datasets containing diverse topics. On the other hand, syntactic models which are content-independent are more robust against topic variance. Zhang et. al. \cite{zhang2018syntax} introduces a strategy to incorporate syntactic information of documents in authorship attribution task. They propose a novel scheme to encode a syntax tree into a learnable distributed representation, and then integrate the syntax representation into a Convolutional Neural Network (CNN)-based model. Different from their approach, we are interested in a neural model which encodes the syntactic information without being equipped with explicit structural representation such as syntax parse tree.  This is achieved by introducing a strategy to encode syntactic information of sentences using only their Part of Speech (POS) tags. 
Furthermore, our motivation is to develop a neural model which preserves all the stylistic information of documents from all three levels of language production including lexical, syntactic, and structural.  

Our contribution in this paper is twofold. First, encoding syntactic information of sentences using only their part of speech tags is more computation efficient and gives better results. Second, we employ a hierarchical neural network to encode the structural information of documents.  This further enhances the performance of the proposed technique. In the proposed model, we use lexical and syntactic embeddings to build two different sentence representations. Subsequently, the lexical and syntactic representations of sentences are independently fed into two parallel hierarchical neural networks to capture semantic and syntactic structure of sentences in documents. The hierarchical attention networks captures the hierarchical structure of documents by constructing representation of sentences and aggregating them into document representations \cite{yang2016hierarchical}. We employ convolutional layers as the word-level encoder to represent each sentence by its important lexical and syntactic n-grams independent of their position in the sentence. For sentence-level encoder, we employ an attention-based recurrent neural network to capture the structural patterns of sentences in the document. The primary reason for adopting recurrent architecture for sentence-encoder is because recurrent neural networks have been shown to be essential for capturing the underlying hierarchical  structure of sequential data \cite{tran2018importance}. Hence, sentence-encoder in the proposed model is expected to capture the structural information of documents. The final document representation is constructed by summing up all the learned sentence vectors while rewarding the sentence which contribute more to the predictions. Ultimately, lexical and syntactic representations are fused and fed into a softmax classifier to predict the probability distribution over the class labels.


\section{Related Work}\label{related_work}
Style-based text classification was introduced by Argamon-Engelson et al. \cite{argamon1998style}. The authors used basic stylistic features (the frequency of function words and part-of-speech trigrams) to classify news documents based on the corresponding publisher (newspaper or magazine) as well as text genre (editorial or news item).  Nowadays, computational stylometry has a wide range of applications in literary science \cite{kabbara2016stylistic, van2017exploring}, forensics \cite{brennan2012adversarial,afroz2012detecting,wang2017liar}, and psycholinguistics \cite{newman2003lying,pennebaker1999linguistic}. 

Syntactic n-grams are shown to achieve promising results in different stylometric tasks including author profiling \cite{posadas2015syntactic} and author verification \cite{krause2014behavioral}. In particular, Raghahvan et al. investigated the use of syntactic information by proposing a probabilistic context-free grammar for the authorship attribution purpose, and used it as a language model for classification \cite{raghavan2010authorship}. A combination of lexical and syntactic features has also been shown to enhance the model performance. Sundararajan et al. argue that, although syntax can be helpful for cross-genre authorship attribution, combining syntax and lexical information can further boost the performance for cross-topic attribution and single-domain attribution \cite{sundararajan2018represents}. Further studies which combine lexical and syntactic features include \cite{ soler2017relevance, schwartz2017effect, kreutz2018exploring}.

With recent advances in deep learning, there exists a large body of work in the literature which employs deep neural networks in authorship attribution domain. For instance,
Ge et al. used a feed forward neural network language model on an authorship attribution task. The output achieves promising results compared to the n-gram baseline \cite{ge2016authorship}. 
Bagnall et al. have employed a recurrent neural network with a shared recurrent state which outperforms other proposed methods in PAN 2015 task \cite{bagnall2016authorship}. 

Shrestha et al. applied CNN based on character n-gram to identify the authors of tweets. Given that each tweet is short in nature, their approach shows that a sequence of character n-grams as the result of CNN allows the architecture to capture the character-level interactions, which can then be aggregated
to learn higher-level patterns for modeling
the style \cite{shrestha2017convolutional}.  
Hitchler et al. propose a CNN based on pre-trained embedding word vector concatenated with one hot encoding of POS tags; however, they have not shown any ablation study to report the contribution of POS tags on the final performance results \cite{hitschler2017authorship}. 
Zhang et.al introduces a syntax encoding approach using convolutional neural networks which combines with a lexical models, and applies it to the domain of authorship attribution \cite{zhang2018syntax}. Their proposed approach utilized syntax parse tree of sentences; however, we show in this paper that such an explicit annotation of hierarchical syntax is not necessary for the authorship attribution task. We propose a simpler yet more effective way of encoding syntactic information of documents for the domain of authorship attribution. Moreover, we employ a hierarchical neural network to capture the structural information of documents and finally introduce a neural model which incorporates all three stylistic features including lexical, syntactic and structural.


\section{Style-aware Neural Model}\label{proposed_method}

We introduce a neural network which encodes the stylistic information of documents from three levels of language production (Lexical, syntactic, and structural). We assume that each document is a sequence of $M$ sentences and each sentence is a sequence of $N$ words, where $M$, and $N$ are model hyperparameters and the best values are explored through the hyperparameter tuning phase (Section \ref{Tuning}).
First, we obtain both lexical and syntactic representation of words using lexical and syntactic embeddings respectively. These two representation are fed into two identical hierarchical neural network which encode the lexical and syntactic patterns of documents independently and in parallel. Ultimately, these two representation are aggregated into the final vector representation of document which is fed into a softmax layer to compute the probability distribution over class labels.

\begin{figure*}[t!]
\centering
\includegraphics[scale=0.33]{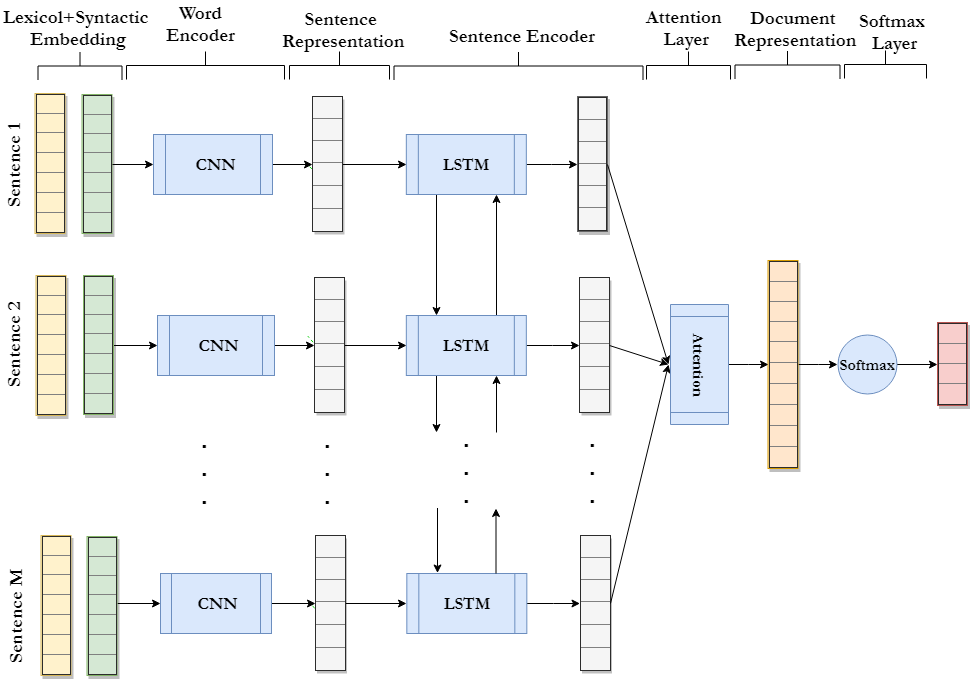}
\caption{The overall architecture of Style-Aware Neural Model}\label{model}
\end{figure*}

The hierarchical neural network is comprised of convolutional layers as word-level encoder to obtain the sentence representations. They are then aggregated into document representation using recurrent neural networks. Finally, we use attention mechanism to reward the sentences which contribute more to the detection of authorial writing style. The overall architecture of the proposed model is shown in figure \ref{model}.  We elaborate each component in the following subsections.


\subsection{Lexical and Syntax Encoding}
\begin{figure}[h!]
\centering
\includegraphics[scale=0.5]{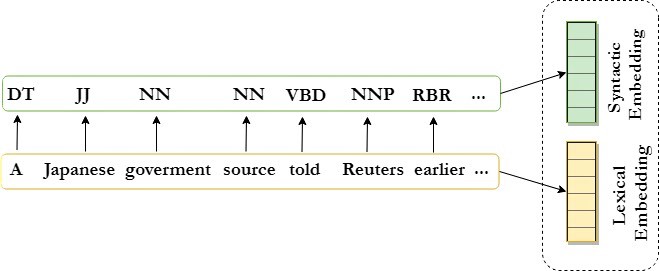}
\caption{Lexical and Syntactic Embedding }\label{jointembedding}
\end{figure}

We encode semantic and syntactic information of documents independently using lexical and syntactic embeddings which is illustrated in figure \ref{jointembedding}. These two representation will fed into two parallel hierarchical networks. Hence the syntactic and semantic patterns of document are learned independently from each other.

\subsubsection{Lexical Embedding}
In lexical-level, we embed each word to a vector representation. We use  pre-trained Glove embeddings \cite{pennington2014glove} and represent each sentence as the sequence of its corresponding word embeddings.

\subsubsection{Syntactic Embedding}
Given a sentence, we convert each word into the corresponding POS tag in the sentence, and then embed each POS tag into a low dimensional vector $P_{i} \in \mathbb{R}^{d_p}$ using a trainable lookup table $ \theta_P \in \mathbb{R}^{|T|\times d_p}$, where $T$ is the set of all possible POS tags in the language. We use NLTK part-of-speech tagger \cite{bird2009natural} for the tagging purpose and use the set of $47$ POS tags\footnote{\url{https://github.com/nltk/nltk/blob/new-corpus-view/nltk/app/chunkparser\_app.py}} in our model as follows.\\

\textit{\small{ T = \{ CC, CD, DT, EX, FW, IN, JJ, JJR, JJS, LS, MD, NN, NNS, NNP, NNPS, PDT, POS, PRP, PRP\$, RB, RBR, RBS, RP, SYM, TO, UH, VB, VBD, VBG, VBN, VBP, VBZ, WDT, WP, WP\$, WRB, 
\lq,\rq, 
\lq:\rq, 
\lq...\rq, 
\lq;\rq, 
\lq?\rq, 
\lq!\rq, 
\lq.\rq, 
\lq\$\rq, 
\lq(\rq, 
\lq)\rq, 
\lq `` \rq, 
\lq '' \rq
\} 
}}\\

One of the advantages of syntax embedding over word embeddings is its low dimensional lookup table compared to the word embeddings, where the size of vocabulary in large datasets usually surpasses 50K words. On the other hand, the size of syntactic embedding lookup table is significantly smaller, fixed, and independent of the dataset which makes the proposed representation less prone to out-of-vocabulary problem.

\subsection{Hierarchical Model}
\subsubsection{Word-level Encoder}
The outputs of lexical and syntactic embedding layer go into two identical convolutional layers (lexical-CNN and Syntactic-CNN) which learn the semantic and syntactic patterns of sentences in parallel. Due to the identical architecture of both networks, we only elaborate on the syntactic-CNN in what follows. 

Let $S_i = [P_1;P_2;...;P_N]$ be the vector representation of sentence $i$, and $W \in \mathbb{R}^{rd_p}$ be the convolutional filter with receptive field size of $r$. We apply a single layer of convolving filters with varying window sizes as the rectified linear unit function (relu) with a bias term b, followed by a temporal max-pooling layer which returns only the maximum value of each feature map $C^{r}_{i} \in \mathbb{R}^{N-r+1}$.
Each sentence is then represented by its most important syntactic n-grams, independent of their position in the sentence.
Variable receptive field sizes $Z$ are used to compute vectors for different n-grams in parallel; and they are concatenated into a final feature vector $h_i \in \mathbb{R}^{K}$ afterwards, where $K$ is the total number of filters:\\
$$ C^{r}_{ij} = relu(W^T S_{j:j+r-1}+b), j \in [1,N-r+1] ,$$
$$ \hat{C^{r}_i} =  max\{C^{r}_i\} , $$
$$ h_i = \oplus \hat{C^{r}_i},     \forall r \in Z $$

\subsubsection{Sentence-level Encoder}

Sentence encoder learns the lexical/syntactic representation of a document from the sequence of sentence representations output from the word-level encoder. We use a bidirectional LSTM To capture how sentences with different syntactic patterns are structured in a document. The vector output from the sentence encoder is calculated as follows.
$$\overrightarrow{h^{d}_{i}} =  LSTM(h^{s}_{i}) , i \in[1, M],$$
$$ \overleftarrow{h^{d}_{i}} = LSTM(h^{s}_{i}) , i \in[M, 1],$$
$$h^{d}_{i} = [\overrightarrow{h^{d}_{i}};\overleftarrow{h^{d}_{i}}]$$


Needless to say, not all sentences are equally informative about the authorial style of a document. Therefore, we incorporate attention mechanism to reward the sentences that contribute more in detecting the writing style. We define a sentence level vector $u_s$ and use it to measure the importance of the sentence $i$ as follows:
$$ u_i = tanh(W_s h^{d}_{i} +b_s)$$
$$ \alpha_i=\frac{exp(u_i^T u_s)}{\sum_i exp(u_i^T u_s)} $$
$$ V = \sum _i \alpha_i h^{d}_{i}$$

Where $u_s$ is a learnable vector and is randomly initialized during the training process and $V$ is the vector representation of document which is weighted sum of vector representations of all sentences.

The primary reason for adopting recurrent architecture for sentence-encoder is because recurrent neural networks have been shown to be essential for capturing the underlying hierarchical structure of sequential data \cite{tran2018importance}. By adopting this approach sentence-encoder is able to encode how sentences are structured in a document. Accordingly, structural information of documents are incorporated into the final document representation.

\subsection{Lexical and Syntactic Representations Fusion }\label{fusion}
In this phase, the semantic and syntactic representations of document learned independently by the two parallel hierarchical neural networks are concatenated into the final vector representation.
$$ {V_k} = [V_{lexical};V_{syntactic}]$$

\subsection{Classification}

The learned vector representation of documents are fed into a softmax classifier to compute the probability distribution of class labels. Suppose $V_k$ is the final vector representation of document $k$ output from the fusion layer. The prediction $\tilde{y_k}$ is the output of softmax layer and is computed as:
$$ \tilde{y_k} = softmax(W_{c} V_{k} + b_c),$$
where $W_c$ and $b_c$ are the learnable weight and learnable bias, respectively; and $\tilde{y_i}$ is a $C$ dimensional vector, where C is the number of classes. We use cross-entropy loss to measure the discrepancy of predictions and the true labels $y_k$. The model parameters are optimized to minimize the cross-entropy loss over all the documents in the training corpus. Hence, the regularized loss function over $X$ documents denoted by $J(\theta)$ is: 
$$ J(\theta) = - \frac{1}{X} \sum_{i=1}^{X} \sum_{k=1}^{C} y_{ik} log \tilde{y}_{ik}+ \lambda ||\theta|| $$

\section{Experimental Studies}\label{experiment_eval}
First, we provide ablation studies to report the contribution of the three stylistic levels (lexical, syntactic, and structural) in the final results. Then we show the performance of our proposed method (Style-HAN) on several benchmark datasets in comparison with the existing baselines in the literature.

\subsection{Datasets}
We evaluate the proposed approach on several benchmark datasets:
\begin{itemize}
\item \textbf{CCAT10 , CCAT50:} Newswire stories from Reuters Corpus Volume 1 (RCV1) written by 10 and 50 authors, respectively \cite{stamatatos2008author}.

\item \textbf{BLOGS10, BLOGS50:} Posts written by 10 and 50 top bloggers respectively, originated from data set of 681,288 blog posts by 19,320 bloggers for blogger.com \cite{schler2006effects}.
\end{itemize}
Some statistics on the sentence length and document length for each dataset are presented in Table \ref{datastat}.

\begin{table}[h!]
\small
\begin{center}
\begin{tabular}{|c|c|c|c|c|}
\hline \bf Param & \bf CCAT10 & \bf CCAT50 & \bf BLOGS10 & \bf BLOGS50  \\ \hline
$|A|$ & 10 & 50 & 10 & 50  \\
S &100 & 100& 874& 682\\ 
W & 580 &584&380 & 331  \\
N & 27 & 26 & 18 & 17  \\
M & 21 & 21& 24& 21 \\

\hline
\end{tabular}
\end{center}
\caption{\label{datastat} Dataset Statistics ($|A|$ : the number of authors, s: the average number of documents per author, w : the average number of words per doc, n: the average number words per sentence, m: the average number of sentences per document)}
\end{table}


\subsection{Baselines}
We compare our method with various baseline approaches which represent the current state of the art in authorship attribution problem, including SVM with affix and punctuation 3-grams \cite{sapkota2015not}, CNN-char \cite{ruder2016character}, Continuous N-gram representation \cite{sari2017continuous}, N-gram CNN \cite{shrestha2017convolutional}, and syntax-CNN \cite{zhang2018syntax}. Their results reported in this paper are obtained from the corresponding papers.

\subsection{Hyperparameter Tuning}\label{Tuning}

The model hyperparameters include the number of sentences per document($M$) and the number of words per sentence($N$), with their best values obtained during the tuning phase. Table \ref{mnhyp} shows the corresponding values for each dataset. The networks are trained using mini-batches with size of 32. We use Nadam optimizer \cite{sutskever2013importance} to optimize the cross entropy loss over 50 epochs of training. We use 100 dimensional pre-trained Glove embeddings \cite{pennington2014glove} for the lexical layer and 100 dimensional randomly initialized embeddings for the syntactic layer. In order to reduce the effect of out-of-vocabulary problem in lexical layer, we retain only 50,000 most frequent words. All the performance metrics are the mean of accuracy (on the test set) calculated over 10 runs with a 0.9/0.1 train/validation split.

\begin{table}[h!]
\begin{center}
\begin{tabular}{|c|c|c|}
\hline \bf Dataset & \bf M & \bf N \\
\hline
CCAT10 & 30 & 40 \\
CCAT50 & 30 & 40 \\
BLOG10 & 20 & 40 \\
BLOG50 & 20 &40 \\
\hline
\end{tabular}
\end{center}
\caption{\label{mnhyp} The Model hyperprameters for each dataset  }
\end{table}

\subsection{Performance Results}

\subsubsection{\bf Syntactic Representation}

First, we compare our proposed syntax encoding method (POS encoding) to the prior method syntax tree (ST) encoding \cite{zhang2018syntax}. In ST encoding, the syntax parse tree of sentences are utilized to encode the syntactic information of sentences. Each word in the sentence is embedded through the corresponding path in the syntax tree. In this approach, the hierarchical structure of sentences are explicitly given as an input to the model. However, we argue if such an explicit annotation is necessary for author attribution. In our proposed POS encoding model, each word is embedded by only its part of speech tag and the neural model itself implicitly learns the dependencies between the parts of speech in the sentences. Furthermore, utilizing only POS tags of words makes the model computationally less expensive when compared to utilizing syntax parse tree structure.

Table \ref{syntax_encoding} reports the accuracy of different syntactic representations for all the benchmark datasets. In ST encoding, the authors uses a CNN-based neural model; hence, we employ the the identical network architecture proposed in the paper in order to have a fair comparison of two different syntactic representations. The results for ST encoding are reported from the corresponding paper. The experimental results demonstrate that our proposed syntactic representation (POS-CNN) outperforms the previously proposed method (ST-CNN) by a large margin in all the benchmark datasets (38.6\% in CCAT10, 30.80\% in CCAT50, 19.62\% in BLOGS10, 11.94\% in BLOGS50).
This improvement in performance can be due two factors. First, the model complexity in POS encoding has been remarkably decreased which makes it more capable of generalization. Second, utilizing syntax tree imposes the positional factor of syntactic units in the sentences. While authorship attribution task is interested to capture the frequent syntactic patterns regardless of their position in the sentences. Our performance results confirm this insight showing that low-level syntax information are more revealing of writing style when compared to hierarchical notion of syntax.

\begin{table}[t!]
\begin{center}
\begin{tabular}{c c c c c }
\hline \bf Model & \bf CCAT10 & \bf CCAT50 & \bf BLOGS10 & \bf BLOGS50 \\ \hline \hline

ST-CNN  & 22.8 & 10.08& 48.64& 42.91 \\
POS-CNN  & 61.40 & 40.98 &  68.26&   54.85  \\ 
POS-HAN & \bf 63.14& \bf 41.30 & \bf 69.32 & \bf 57.76 \\ 

\hline
\end{tabular}
\end{center}
\caption{\label{syntax_encoding} The Accuracy of Different Syntactic Representations}
\end{table}

\subsubsection{\bf Hierarchical Neural Model}
We have employed hierarchical attention network (HAN) in order to capture the structural information of documents. In order to understand the contribution of our network architecture to the performance, we compare our network architecture (POS-HAN) to the previously proposed CNN-based model (POS-CNN) when the syntactic representations are kept identical. According to Table \ref{syntax_encoding}, POS-HAN outperforms POS-CNN model consistently across all the benchmark datasets (1.74\% in CCAT10, 0.32\% in CCAT50, 1.06\% in BLOGS10, 2.91\% in BLOGS50). This observation indicates that hierarchical neural models which capture the hierarchical structure of documents are a better choice for authorship attribution task.  This confirms our argument that structural information of the document is important to reveal the authorial writing style. 

\subsubsection{\bf Lexical and Syntactic Model}

In order to understand the contribution of lexical and syntactic models to the final predictions, we performed an ablation study.  The results are reported in table \ref{ablationstudy}.
In Syntactic-HAN, only syntactic representation of documents ($V_{syntactic}$) is fed into the softmax layer to compute the final predictions. Similarly, in Lexical-HAN, only lexical representation of documents ($V_{lexical}$) is fed into the softmax classifier. The final stylometry model, Style-HAN, fuses both representations and computes the class labels using a softmax classifier (Section \ref{fusion}). According to the table, lexical model consistently outperforms the syntactic-model across all the benchmark datasets. Moreover, combining the two representations further improves the performance results.

\begin{table}[h!]
\begin{center}
\begin{tabular}{c c c c c }
\hline \bf Model & \bf CCAT10 & \bf CCAT50 & \bf BLOGS10 & \bf BLOGS50 \\ \hline \hline

Syntactic-HAN & 63.14& 41.30 & 69.32 & 57.76 \\ 
Lexical-HAN &86.04 & 79.50 & 70.81 &  59.77\\ 
Style-HAN & \bf 90.58 & \bf 82.35 & \bf 72.83 & \bf 61.19 \\ 

\hline
\end{tabular}
\end{center}
\caption{\label{ablationstudy} The Accuracy of Syntactic (Syntactic-HAN), Lexical (Lexical-HAN),and combined (Style-HAN) Models}
\end{table}

\begin{figure*}[h!]
\centering
\subfigure[]{\includegraphics[scale=0.44]{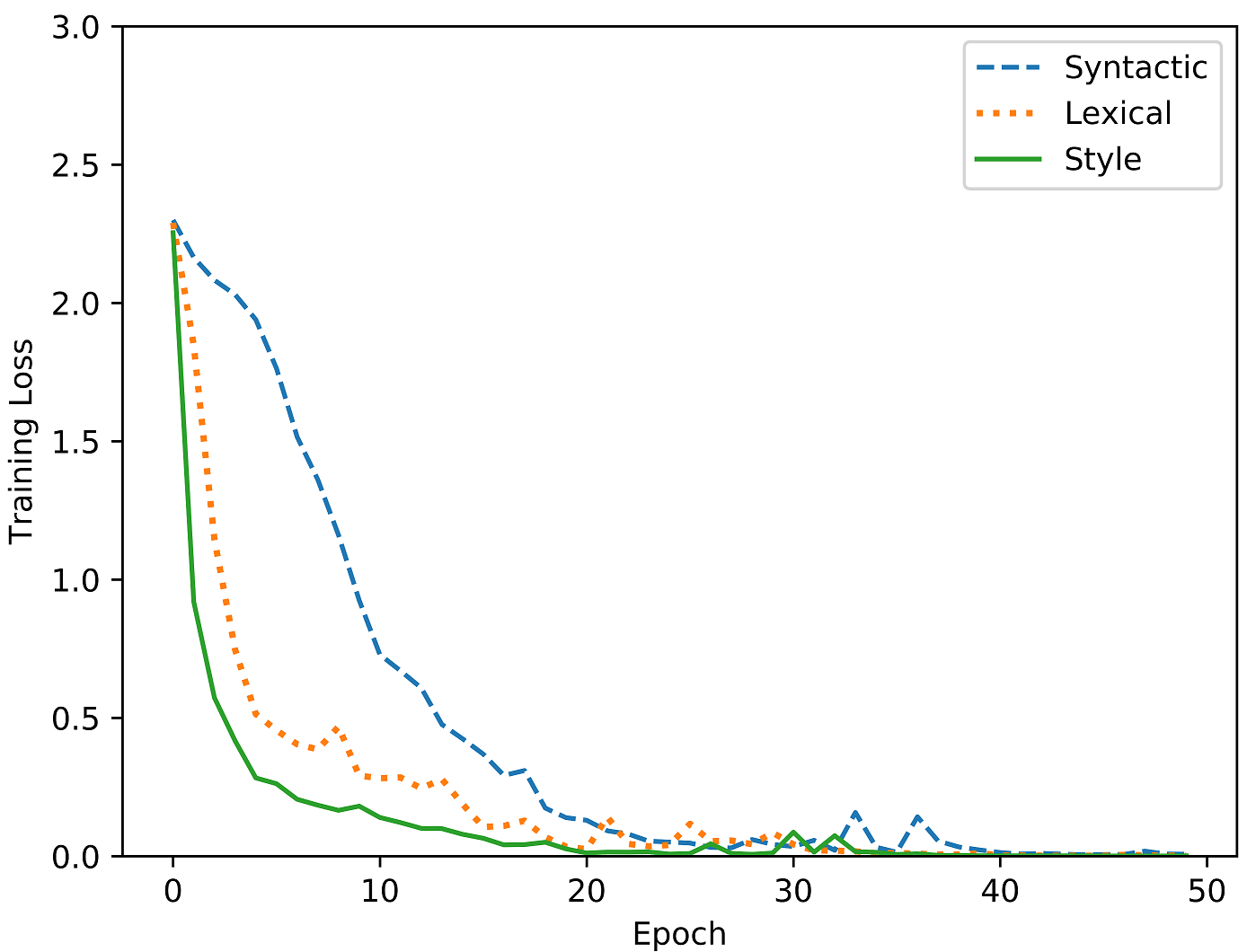}}
\subfigure[]{\includegraphics[scale=0.44]{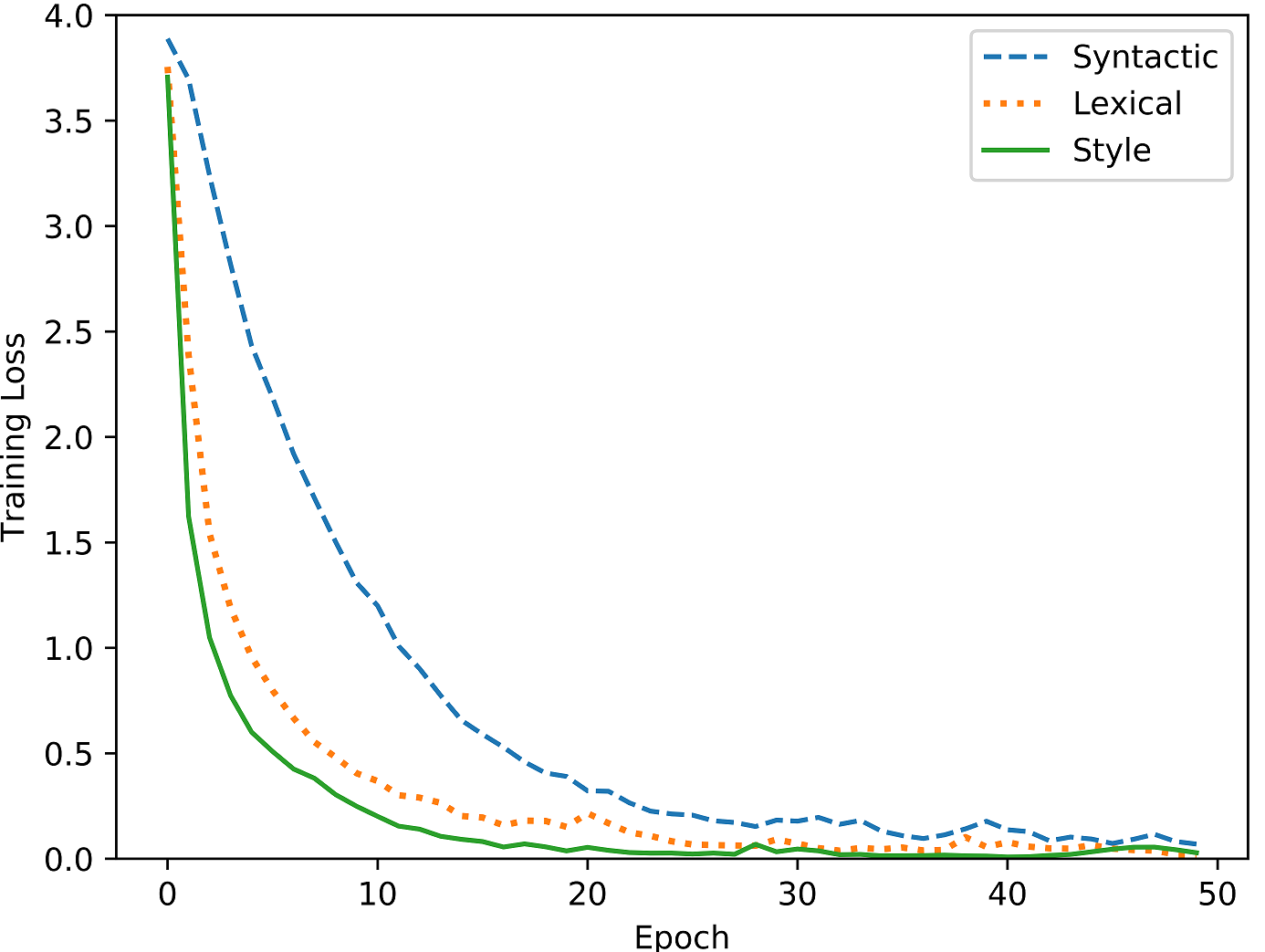}}
\subfigure[]{\includegraphics[scale=0.44]{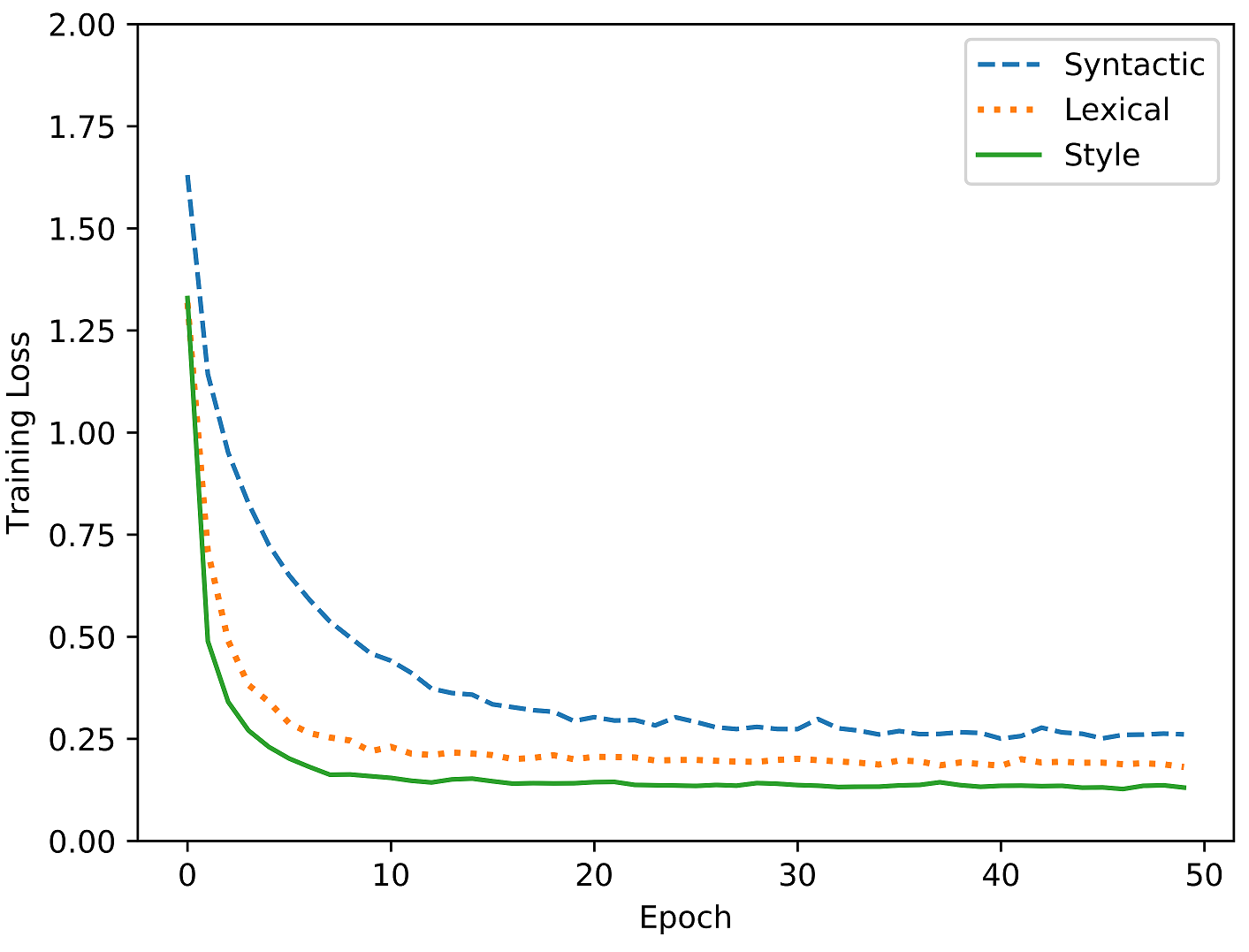}}
\subfigure[]{\includegraphics[scale=0.44]{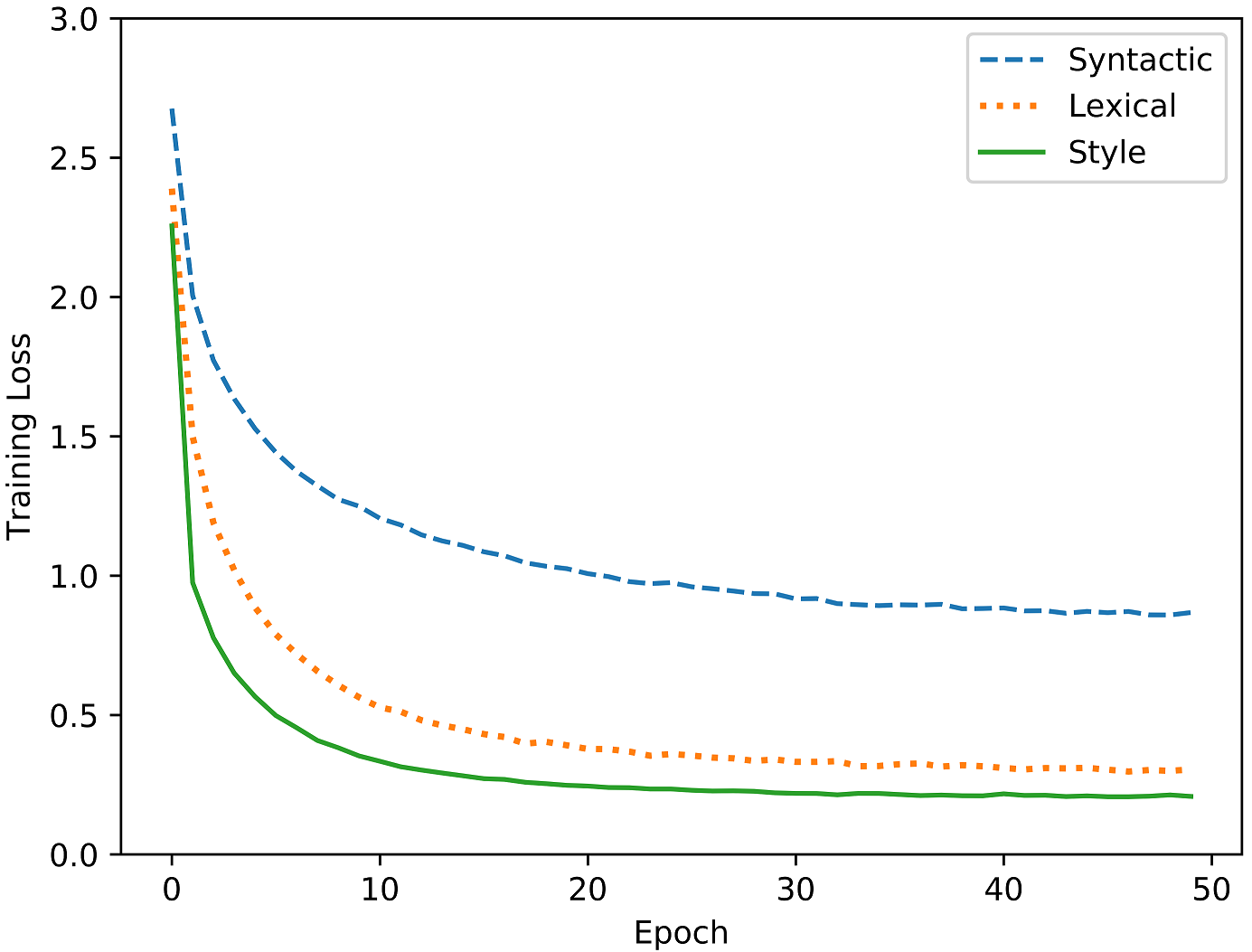}}

\caption{Cross Entropy loss over 50 epochs of training for syntactic, lexical, and style models for (a) CCAT10, (b) CCAT50 , (c) BLOGS10 , and (d) BLOGS50 datasets}  \label{trainingloss}
\end{figure*}

Figure \ref{trainingloss} illustrates the training loss of the syntax, lexical, and style encoding over 50 epochs of training for all the datasets.  As we observe, the lexical model maintains lower loss in the earlier epochs and converges faster when compared to the syntactic model. However, combining them into the style model reduces the loss and improves the performance. 

Based on the observation from figure \ref{trainingloss} and table \ref{ablationstudy}, we realize that even if Syntactic-HAN achieves a comparable performance results combining it with Lexical-HAN, slightly improves the overall performance (Style-HAN). This can be due to the fact that lexical-based recurrent neural networks alone are able to encode significant amount of syntax even in the absence of explicit syntactic annotations \cite{blevins2018deep}. However, explicit syntactic annotation further improves the performance results when it's  compared to lexical-based model. As shown in Table \ref{ablationstudy}, the performance improvement in terms of accuracy is consistent across all the benchmark datasets (4.54\% in CCAT10, 2.85\% in CCAT50, 2.02\% in BLOGS10, 1.42\% in BLOGS50)

\subsubsection{\bf Training Syntactic and Lexical Networks}

\begin{figure}[h]
\centering
\includegraphics[scale=0.3]{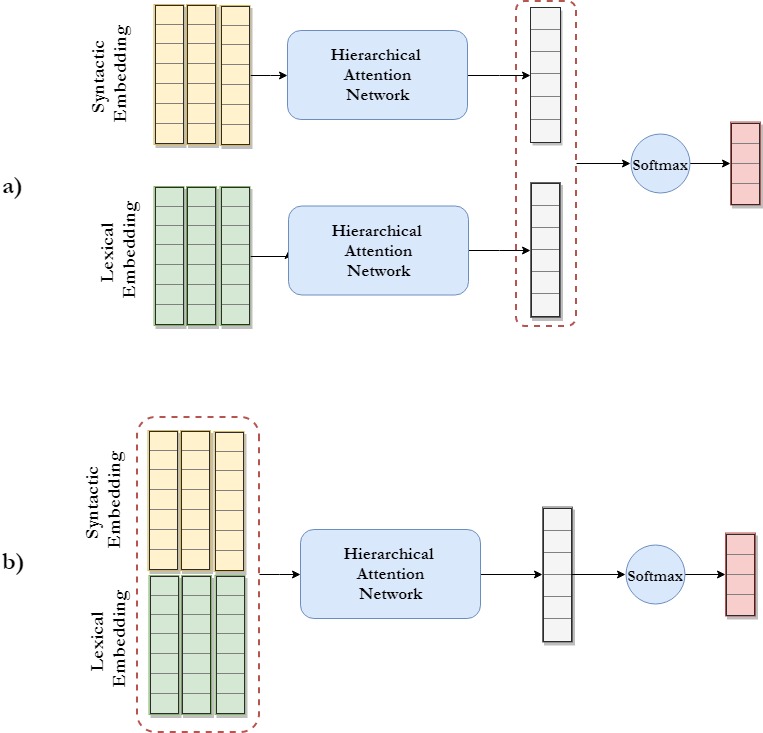}
\caption{Parallel a) and combined b) training of syntactic and lexical representations }\label{architectures}
\end{figure}

We examine two different approaches (combined and parallel) for fusing lexical and syntactic encoding into the final style network. In the combined approach, we concatenate the syntactic and lexical embeddings and construct a unified representation for each word which contains both lexical and syntactic information. Subsequently, this representation is fed to a hierarchical attention network to learn the final document representation.  In the parallel approach, the lexical and syntactic embeddings are fed into two identical hierarchical neural networks and the syntactic and lexical representations of documents which are learned independently and in parallel are concatenated into a final document representation. Figure \ref{architectures} illustrates these two approaches. 
Table \ref{architectures_table} reports the accuracy of the combined and the parallel fusion approaches. According to these results, training two parallel networks for lexical and syntax encoding achieves higher accuracy when compared to training the same network with combined embeddings. This observation can be due to the fact that syntactic and lexical models contain almost complementary information which are language structure and semantics, respectively and training them independently delivers better results.

\begin{table}[h!]
\begin{center}
\begin{tabular}{|c|c|c|}
\hline \bf Dataset & \bf Combined & \bf Parallel \\
\hline
CCAT10 & 88.36 & 90.58 \\
CCAT50 & 81.21 & 82.35 \\
BLOG10 & 67.38 & 72.83 \\
BLOG50 & 58.81 & 61.19 \\
\hline
\end{tabular}
\end{center}
\caption{\label{architectures_table} The accuracy of Different Fusion Approaches  }
\end{table}

\subsubsection{\bf Style Encoding}

\begin{table*}[t!]
\begin{center}
\begin{tabular}{c c c c c }
\hline \bf Model & \bf CCAT10 & \bf CCAT50 & \bf BLOGS10 & \bf BLOGS50 \\ \hline \hline
SVM-affix-punctuation 3-grams & 78.8 & 69.3 & \# & \#   \\
CNN-char & \# & \# & 61.2 & 49.4  \\
Continuous n-gram & 74.8 & 72.6 & 61.34 & 52.82   \\
N-gram CNN & 86.8& 76.5& 63.74& 53.09  \\

Syntax-CNN & 88.20& 81.00& 64.10 & 56.73 \\

Style-HAN & \bf 90.58 & \bf 82.35 & \bf 72.83 & \bf 61.19 \\ 

\hline
\end{tabular}
\end{center}
\caption{\label{results} Test Accuracy of models for each dataset }
\end{table*}

We compare our proposed style-aware neural model (Style-HAN) with the other stylometric models in the literature. Table \ref{results} reports the accuracy of the models on the four benchmark datasets. All the results are obtained from the corresponding papers, with the dataset configuration kept identical for the sake of fair comparison. The best performance result for each dataset is highlighted in bold. It shows that Style-HAN outperforms the baselines by 2.38\%, 1.35\%, 8.73\%, and 4.46\% over the CCAT10, CCAT50, BLOGs10, and BLOGS50 datasets, respectively. This indicates the effectiveness of encoding document information in three stylistic levels including lexical, syntactic and structural.

\subsubsection{\bf Sensitivity to Sentence Length}
\begin{figure}[h!]
\centering
\includegraphics[scale=0.67]{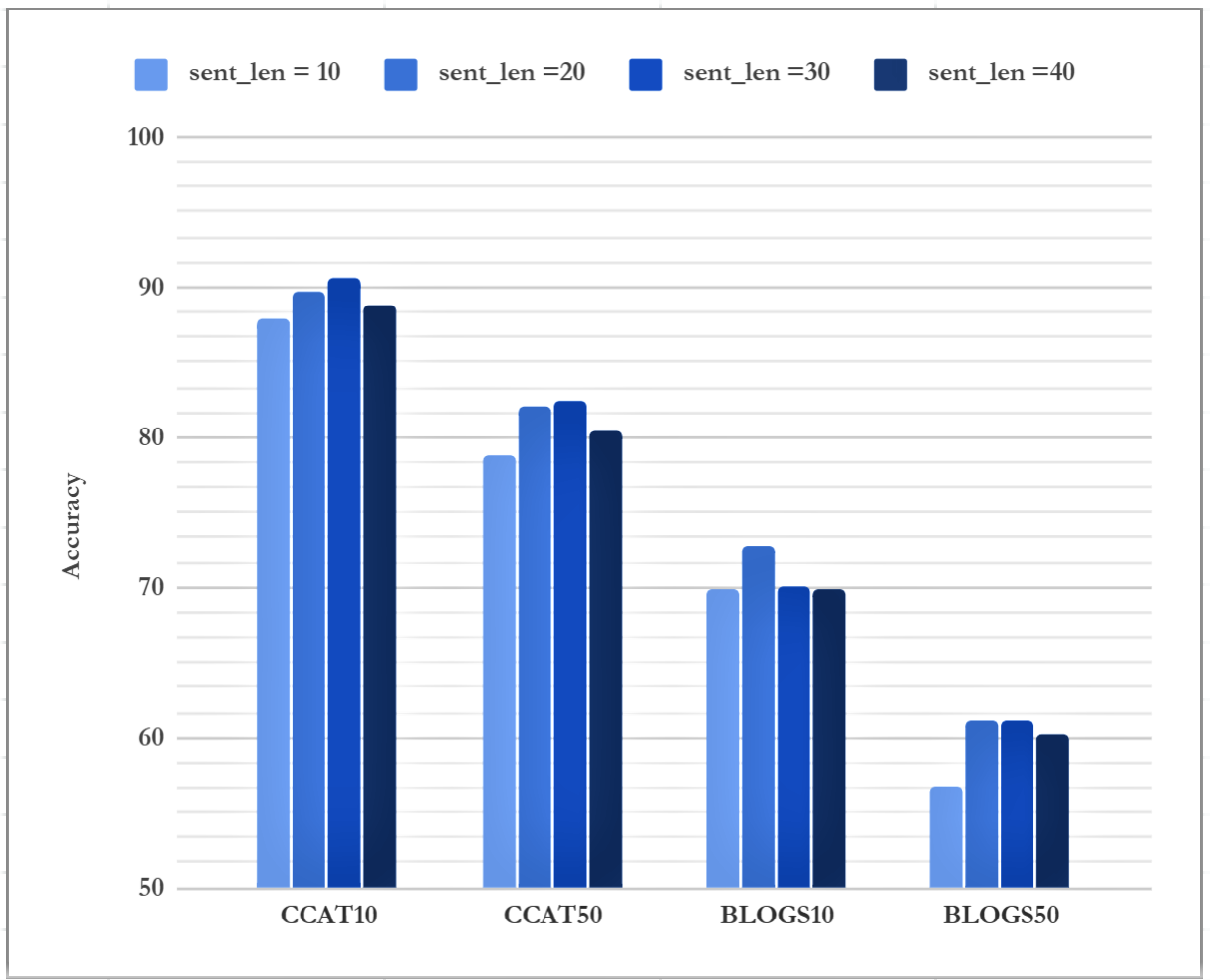}
\caption{The accuracy of Style-HAN model across different sentence length}\label{sents}
\end{figure}

We examine our model's sensitivity to the sentence length (M). We evaluate the performance of the model on different sentence lengths of 10,20,30,and 40 words while the sequence length (N) is kept constant. Figure \ref{sents} shows the performance results of the Style-HAN on the four datasets. It shows that the model achieves the best performance in CCAT10 and CCAT50 when the sentence length is equal to 30; while in BLOGS10 and BLOGS50, the highest performance is observed when the sentence length is equal to 20.  Table \ref{datastat} shows the average sentence lengths in all samples in CCAT10, CCAT50, BLOGS10, and BLOGS50 are 27, 26, 18, and 17, respectively. Accordingly, the models perform better when the sentence length is close to the average sentence length in the dataset. This is simply because a shorter sentence length results in information loss and on the other hand a higher sentence length leads to capturing misleading features.  Both situations result in lower accuracy.

\subsubsection{\bf Sensitivity to Document Length}
\begin{figure}[h!]
\centering
\includegraphics[scale=0.67]{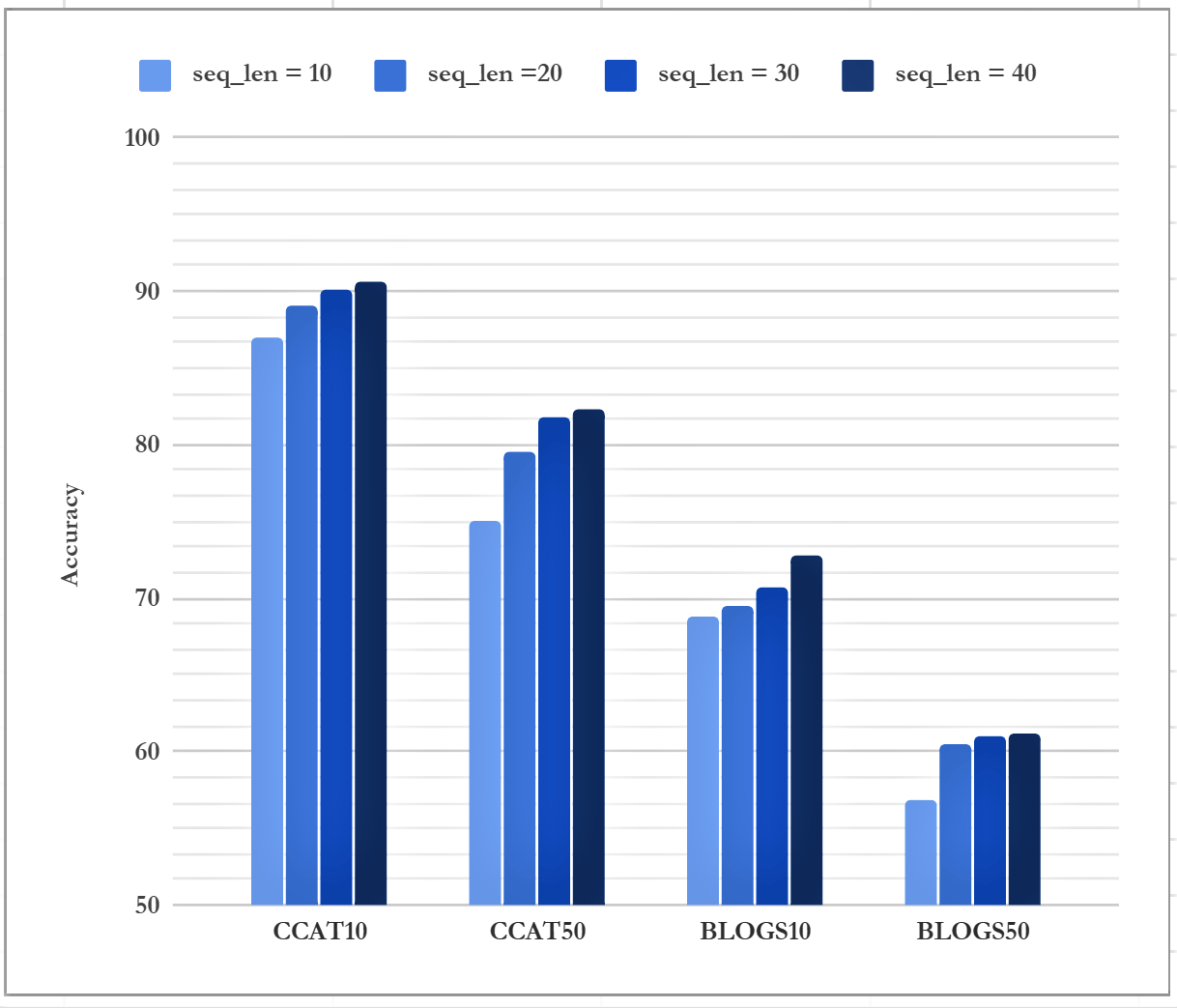}
\caption{The accuracy of Style-HAN model across different document length}\label{seqences}
\end{figure}

We examine the model performance across different number of sentences per document(document length). Figure \ref{seqences} illustrates the accuracy of the model when the number of sentences per document is assumed to be 10, 20, 30, and 40, respectively. We observe that increasing the sequence length (the number of sentences in document) generally boosts the performance on all the datasets.
This observation confirms the fact that investigation of writing style in short documents is more challenging \cite{neal2017surveying}.

\section{Conclusion}\label{conclusions}
In this paper we introduce a style-aware neural model which encodes document information from three stylistic levels including lexical, syntactic, and structural in order to better capture the authorial writing style. First, we propose an efficient way to encode the syntactic patterns of sentences using only their corresponding part-of-speech tags. Lexical and syntactic embeddings of words are then used to create two different sentence representations. Subsequently, a hierarchical neural network is employed to capture the structural patterns of sentences in the document, which takes both syntactic and lexical information as input. Finally, these syntactic and lexical representation of documents are concatenated in the fusion step to build the final document representation. Our experimental results on the benchmark datasets in authorship attribution tasks confirm the benefits of encoding document information from all three stylistic levels, and show the performance advantages of our techniques.


\bibliographystyle{IEEEtran}
\bibliography{references}

\end{document}